\documentclass[runningheads]{llncs}
\usepackage{graphicx}
\usepackage{tikz}
\usepackage{comment}
\usepackage{amsmath,amssymb} % define this before the line numbering.
\usepackage{color}
\usepackage{footmisc}

\usepackage[accsupp]{axessibility}  % Improves PDF readability for those with disabilities.

\usepackage{multirow}
\usepackage[pagebackref=true,breaklinks=true,colorlinks,bookmarks=false]{hyperref}
\usepackage{arydshln} % \hdashline
\usepackage{pdfrender}
\newcommand*{\boldcheckmark}{%
	\textpdfrender{
		TextRenderingMode=FillStroke,
		LineWidth=.5pt, % half of the line width is outside the normal glyph
	}{\checkmark}%
}
\usepackage{wrapfig}

\definecolor{mygreen}{rgb}{0.09, 0.45, 0.27}

\newcommand{\ie}{\textit{i}.\textit{e}.}
\newcommand{\eg}{\textit{e}.\textit{g}.}
\newcommand{\cf}{\textit{cf.}}
\newcommand{\etal}{\textit{et}.\textit{al}.}

\newcommand{\aka}{\textit{a}.\textit{k}.\textit{a}.}

\newcommand{\Qyesno}{``\texttt{Yes/No}"}
\newcommand{\Qcolor}{``\texttt{Color}"}
\newcommand{\Qnumber}{``\texttt{Number}"}
\newcommand{\Qwhat}{``\texttt{What}"}

\begin{document}
% \linenumbers
\pagestyle{headings}
\mainmatter
\def\ECCVSubNumber{2237}  % Insert your submission number here

% \title{Rethinking Data Augmentation for Robust Visual Question Answering: A New Framework} % Replace with your title
\title{Rethinking Data Augmentation for Robust Visual Question Answering}

% \begin{comment}
% \titlerunning{ECCV-22 submission ID \ECCVSubNumber} 
% \authorrunning{ECCV-22 submission ID \ECCVSubNumber} 
% \author{Anonymous ECCV submission}
% \institute{Paper ID \ECCVSubNumber}
% \end{comment}
%******************

% \begin{comment}
\titlerunning{Rethinking Data Augmentation for Robust Visual Question Answering}
% \author{Long Chen\inst{1}\orcidID{0000-1111-2222-3333} \and
% Yuhang Zheng\inst{2}\orcidID{1111-2222-3333-4444} \and
% Jun Xiao\inst{2}\orcidID{2222--3333-4444-5555}}
\author{Long Chen\inst{1}\thanks{Long Chen and Yuhang Zheng are co-first authors with equal contributions.} \and
Yuhang Zheng\inst{2}$^\ast$ \and
Jun Xiao\inst{2}\thanks{Corresponding author. Codes: \url{https://github.com/ItemZheng/KDDAug}.}}
% \orcidlink{0000-0001-6148-9709}
% \orcidlink{0000-0001-9628-1940}
%
\authorrunning{L. Chen and Y. Zheng et al.}
\institute{$^1$Columbia University \qquad
$^2$Zhejiang University \\
\email{zjuchenlong@gmail.com, itemzhang@zju.edu.cn, junx@cs.zju.edu.cn}
% \url{https://github.com/ItemZheng/KDDAug}
}
% \institute{Columbia University \\
% \and
% Zhejiang University} \\

% \email{zjuchenlong.com} \\
% % \and

% \email{\{itemzhang,junx\}@zju.edu.cn}}
% \end{comment}
%******************
\maketitle

\begin{abstract}
Data Augmentation (DA) --- generating extra training samples beyond the original training set --- has been widely-used in today's unbiased VQA models to mitigate language biases. Current mainstream DA strategies are synthetic-based methods, which synthesize new samples by either editing some visual regions/words, or re-generating them from scratch. However, these synthetic samples are always unnatural and error-prone. To avoid this issue, a recent DA work composes new augmented samples by randomly pairing pristine images and other human-written questions. Unfortunately, to guarantee augmented samples have reasonable ground-truth answers, they manually design a set of heuristic rules for several question types, which extremely limits its generalization abilities. To this end, we propose a new \textbf{K}nowledge \textbf{D}istillation based \textbf{D}ata \textbf{Aug}mentation for VQA, dubbed \textbf{KDDAug}. Specifically, we first relax the requirements of reasonable image-question pairs, which can be easily applied to any question type. Then, we design a knowledge distillation (KD) based answer assignment to generate pseudo answers for all composed image-question pairs, which are robust to both \emph{in-domain} and \emph{out-of-distribution} settings. Since KDDAug is a model-agnostic DA strategy, it can be seamlessly incorporated into any VQA architecture. Extensive ablation studies on multiple backbones and benchmarks have demonstrated the effectiveness and generalization abilities of KDDAug.


\end{abstract}

\section{Introduction} \label{sec:1}

Visual Question Answering (\textbf{VQA}), \ie, answering any natural language questions about the given visual content, is regarded as the holy grail of a human-like vision system~\cite{geman2015visual}. Due to its multi-modal nature, VQA has raised unprecedented attention from both CV and NLP communities, and hundreds of VQA models have been developed in recent years. Although current VQA models can achieve really ``decent" performance on standard benchmarks, numerous studies have revealed that today's models tend to over-rely on the superficial linguistic correlations between the questions and answers rather than multi-modal reasoning (\aka, \textbf{language biases})~\cite{agrawal2018don,agrawal2016analyzing,zhang2016yin,johnson2017clevr,goyal2017making}. For example, blindly answering ``2" for all counting questions or ``tennis" for all sport-related questions can still get a satisfactory performance. To mitigate these bias issues and realize robust VQA, a recent surge of VQA work~\cite{chen2020counterfactual,chen2021counterfactual,kolling2022efficient,agarwal2020towards,gokhale2020mutant,boukhers2022coin,tang2020semantic,kant2021contrast,kafle2017data,gokhale2020vqa,bitton2021automatic,narjes2022inductive,wang2021cross} resort to different data augmentation techniques (\ie, generating extra training samples beyond original training set), and achieve good performance on both the in-domain (ID) (\eg, VQA v2~\cite{goyal2017making}) and out-of-distribution (OOD) datasets (\eg, VQA-CP~\cite{agrawal2018don}).

% \footnote{The accuracy of SOTA VQA models on benchmark VQA v2~\cite{goyal2017making} has surpassed 80\%.}

\begin{figure}[t]
	\centering
	\setlength{\abovecaptionskip}{-0.5em}
	\setlength{\belowcaptionskip}{-1em}
	\includegraphics[width=\linewidth]{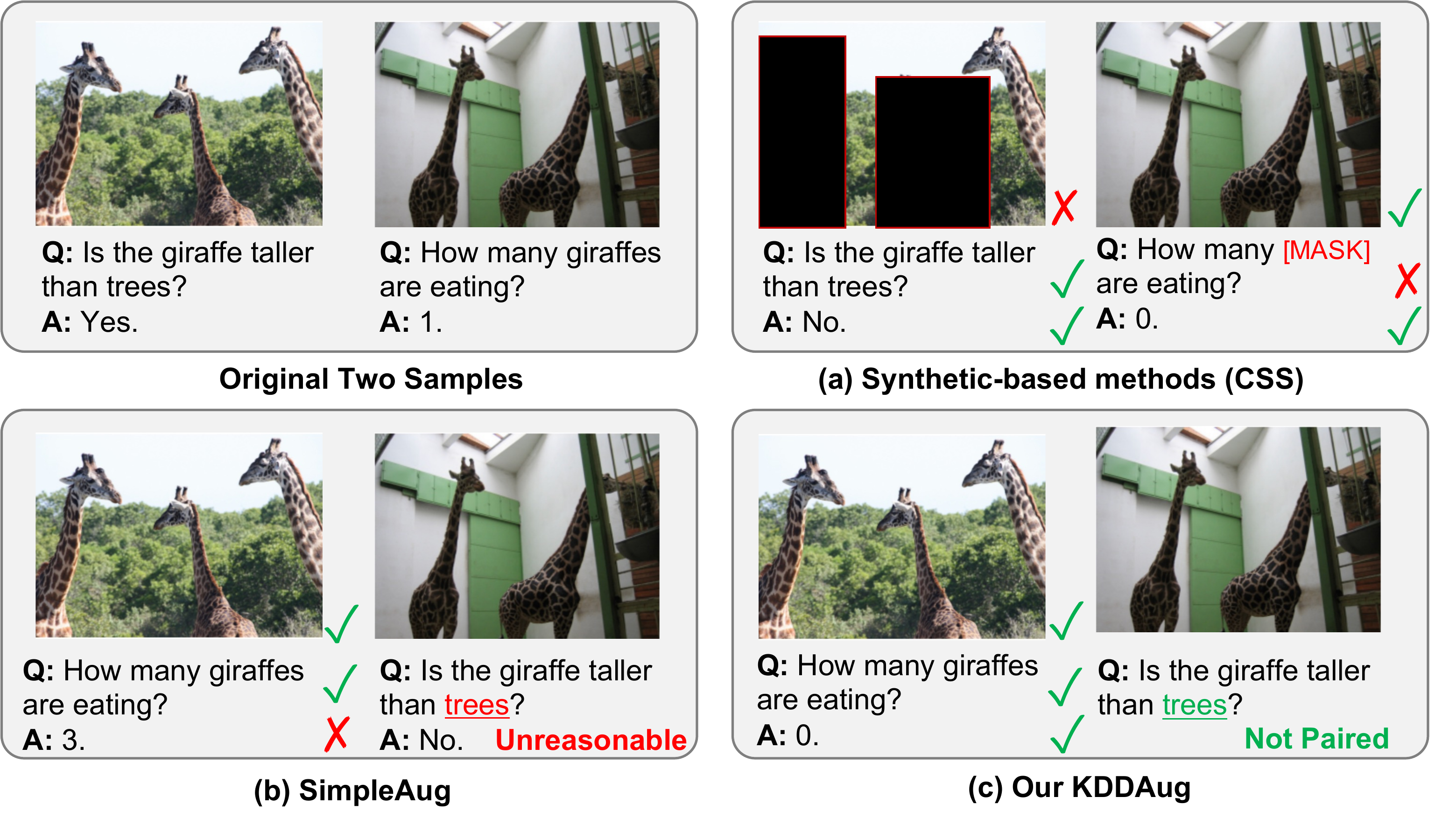}
	\caption{Comparisons between different DA methods for VQA. (a) \textbf{Synthetic-based methods}: Take CSS~\cite{chen2020counterfactual} as an example, it masks some regions or words in the original samples. (b) \textbf{SimpleAug}~\cite{kil2021discovering}: It pairs images with some specific types of questions, and obtains pseudo labels by predefined heuristic rules. (c) \textbf{KDDAug}: It pairs image with any types of questions, and use a KD-based model to predict pseudo answers. Question ``\texttt{is \dots}" is not a reasonable question for the right image as it contains ``\texttt{trees}".}
	\label{fig:motivation_comparisons}
\end{figure}

Currently, mainstream Data Augmentation (DA) strategies for robust VQA are \textbf{synthetic-based} methods. As shown in Fig.~\ref{fig:motivation_comparisons}(a), from modality viewpoint, these synthetic-based DA methods can be categorized into two groups: 1) \emph{Visual-manipulated}: They usually edit some visual regions in original images~\cite{chen2020counterfactual,chen2021counterfactual,kolling2022efficient}, or re-generate counterfactual/adversarial images with generative networks~\cite{agarwal2020towards,gokhale2020mutant} or adversarial attacks~\cite{tang2020semantic}. 2) \emph{Textual-generated}: They edit some words in original questions~\cite{chen2020counterfactual,chen2021counterfactual,kolling2022efficient} or re-generate the sentence from scratch with back translation~\cite{tang2020semantic,kant2021contrast} or visual question generation (VQG) methods~\cite{wang2021cross}. Although these synthetic-based methods have dominated the performance on OOD benchmarks (\eg, VQA-CP), and significantly improved VQA models' interpretability and consistency, there are several inherent weaknesses: 1) Photo-realistic image generation or accurate sentence generation themselves are still open problems, \eg, a significant portion of the generated questions have grammatical errors~\cite{tang2020semantic}. 2) They always need extra human annotations to assign reasonable answers~\cite{gokhale2020mutant}.

To avoid these unnatural and synthetic training samples, a recent work SimpleAug~\cite{kil2021discovering} starts to compose new training samples by randomly pairing images and questions. As the example shown in Fig.~\ref{fig:motivation_comparisons}(b), they pair the left image and human-written questions from the right image (\ie, ``\texttt{How many giraffes are eating?}") into a new image-question (VQ) pair. To obtain ``reasonable" pseudo ground-truth answers for these new VQ pairs, they manually design a set of heuristic rules for several specific question types, including \Qyesno, \Qcolor, \Qnumber, and \Qwhat~type questions. Although SimpleAug avoids the challenging image/sentence generation procedures, there are still several drawbacks: 1) These predefined rules for answer assignment are fallible (\eg, the human-check accuracy for \Qyesno~type questions is only 52.20\%, slightly higher than a random guess.) 2) Due to the limitations of these rules, it only covers several specific types of answers, and it is difficult to extend to other question types\footnote{VQA datasets typically have much more question types (\eg, 65 for VQA v2~\cite{goyal2017making}).}. 3) It still relies on some human annotations (\eg, object annotations in COCO).

\begin{wrapfigure}{r}{0.46\linewidth}
	\centering
	\setlength{\abovecaptionskip}{0em}
	\setlength{\belowcaptionskip}{-2em}
    \includegraphics[width=0.98\linewidth]{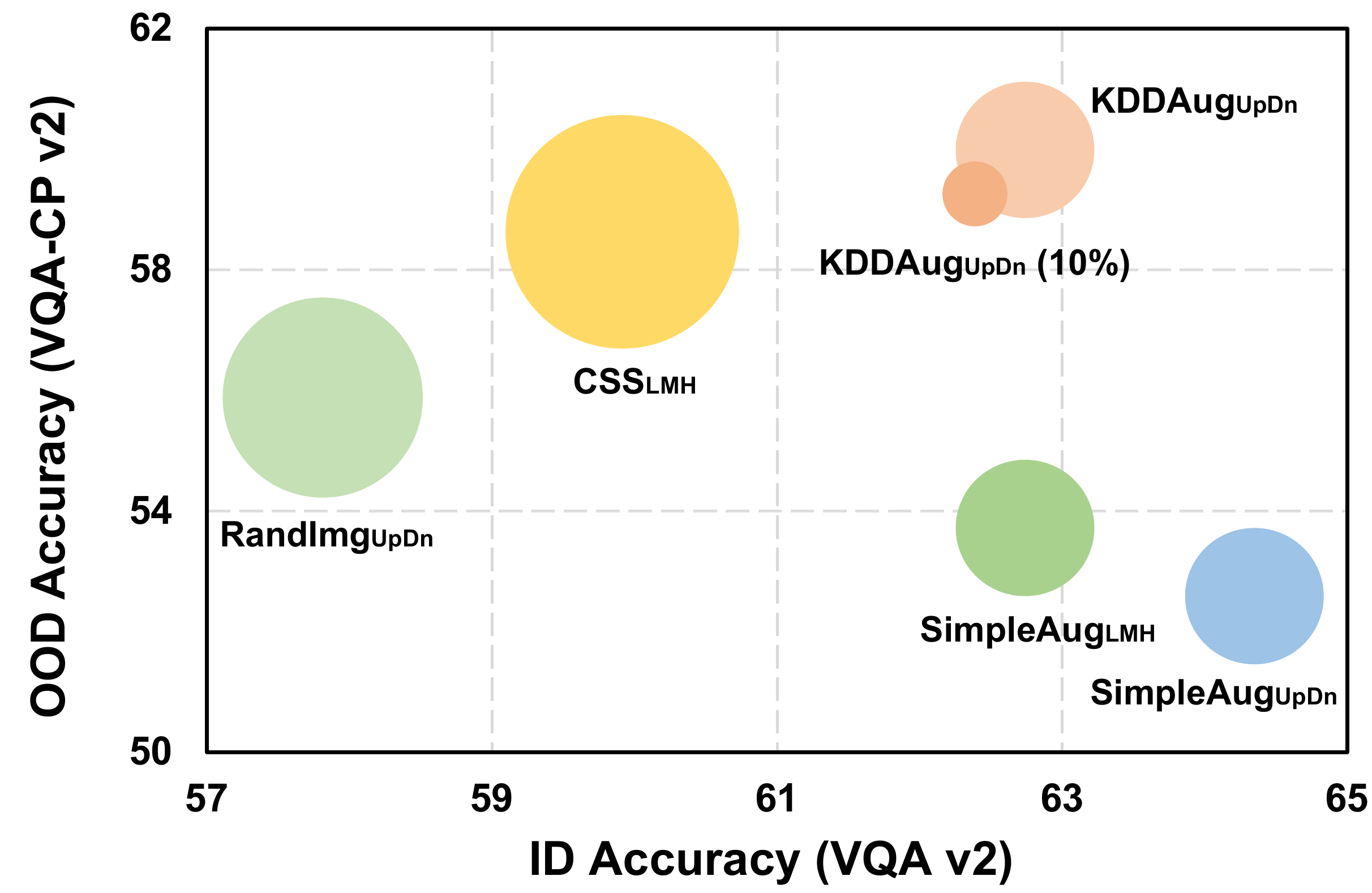}
	\caption{Performance of SOTA DA methods. Circle sizes are in proportion to the number of their augmented samples\footref{foot:fig2}. % KDDAug can achieve the best trade-off performance with even fewer samples.
	}
	\label{fig:id_vs_ood}
\end{wrapfigure}

In this paper, we propose a \textbf{K}nowledge \textbf{D}istillation based \textbf{D}ata \textbf{Aug}mentation (\textbf{KDDAug}) strategy for robust VQA, which can avoid all the mentioned weaknesses in existing DA methods. Specifically, we first relax the requirement for reasonable VQ pairs by only considering the object categories in the images and nouns in the questions. As illustrated in Fig.~\ref{fig:motivation_comparisons}(c), question ``\texttt{how many \underline{giraffes} are eating}" is a reasonable question for the left image which contains ``\texttt{giraffe}" objects. To avoid extra human annotations, we only utilize an off-the-shelf object detector to detect objects\footnote{Since all the compared state-of-the-art VQA models follow UpDn~\cite{anderson2018bottom} and use VG~\cite{krishna2017visual} pretrained detector to extract visual features, we don't use extra annotations.}. After obtaining all the reasonable VQ pairs, we design a multi-teacher knowledge distillation (KD) based answer assignment to generate corresponding pseudo ground-truth answers. We first pretrain two teacher models (ID and OOD teacher) with the original training set, and then utilize these teacher models to predict a ``soft" answer distribution for each VQ pair. Last, we combine the predicted distributions (\ie, knowledge) from both teachers, and treat them as the pseudo answers for augmented VQ pairs. Benefiting from our designs, KDDAug achieves the best trade-off results on both ID and OOD settings with even fewer samples (Fig.~\ref{fig:id_vs_ood})\footnote{CSS \& RandImg dynamically generate different samples in each epoch, their sizes are difficult to determine. Here, their sizes are for illustration (larger than SimpleAug). \label{foot:fig2}}.

Extensive ablation studies have demonstrated the effectiveness of KDDAug. KDDAug can be seamlessly incorporated into any VQA architecture, and consistently boost their performance. Particularly, by building on top of some SOTA debiasing methods (\eg, LMH~\cite{clark2019don}, RUBi~\cite{cadene2019rubi}, and CSS~\cite{chen2020counterfactual}), KDDAug consistently boost their performance on both ID and OOD benchmarks. 

In summary, we make three main contributions in this paper:

\begin{enumerate}

    \item We systematically analyze existing DA strategies for robust VQA, and propose a new KDDAug that can avoid all the weaknesses of existing solutions.
    
    \item We use multi-teacher KD to generate pseudo answers, which not only avoids human annotations, but also is more robust to both ID and OOD settings.
    
    \item KDDAug is a model-agnostic DA method, which empirically boosts multiple different VQA architectures to achieve state-of-the-art performance. 
\end{enumerate}

\section{Related Work} 

\noindent\textbf{Language Biases in VQA.} In order to overcome the language biases issues in VQA, many debiasing methods have been proposed recently. Specifically, existing methods can be roughly divided into two groups: 1) Ensemble-based debiasing methods. These methods always design an auxiliary branch to explicit model and exclude the language biases~\cite{ramakrishnan2018overcoming,grand2019adversarial,cadene2019rubi,clark2019don,rabeeh2020end,niu2021counterfactual,han2021greedy,liang2021lpf,wen2021debiased}. 2) Model-agnostic debiasing methods. These methods mainly include balancing datasets~\cite{goyal2017making,zhang2016yin}, data augmentation by generating augmented training samples~\cite{gokhale2020mutant,chen2020counterfactual,chen2021counterfactual,kil2021discovering,abbasnejad2020counterfactual,kant2021contrast}, and designing extra training objectives~\cite{gokhale2020mutant,zhu2020overcoming,liang2020learning}. Almost existing debiasing methods significantly improve their OOD performance, but with the cost of ID performance drops. In this paper, we deeply analyze existing DA methods, and propose a new DA strategy to achieve a decent trade-off between ID \& OOD performance.

\noindent\textbf{Data Augmentation in VQA.} In addition to the mainstream synthetic-based methods, there are other DA methods: some existing methods generate negative samples by randomly selecting images or questions~\cite{teney2020value,zhu2020overcoming}, or compose reasonable image-question (VQ) pairs as new positive training samples~\cite{kil2021discovering}. For these generated VQ pairs, they utilize manually pre-defined rules to obtain answers, which are designed for some specific question types. However, these DA methods almost either suffer a severe ID performance drop~\cite{chen2020counterfactual,teney2020value,chen2021counterfactual,kolling2022efficient} or their answer assignment mechanisms rely on human annotations and lack generality~\cite{kafle2017data,gokhale2020mutant,gokhale2020vqa,kil2021discovering,narjes2022inductive}. Instead, our KDDAug overcomes all these weaknesses.

\noindent\textbf{Knowledge Distillation.} KD is a method that helps the training process of a smaller student network under the supervision of a larger teacher network~\cite{wang2021knowledge}. The idea of KD has been applied to numerous vision tasks, \eg, object detection~\cite{wang2019distilling,chen2017learning} or visual-language tasks~\cite{zhang2020object,pan2020spatio,li2022integrating}. Recently, Niu~\etal~\cite{niu2021introspective} began to study KD for VQA and propose a KD-based method to generate ``soft" labels in training. Inspired by them, we propose to use a multi-teacher KD to generate robust pseudo ground-truth labels for all new composed VQ pairs.

\section{KDDAug: A New DA Framework for VQA}

Following same conventions of existing VQA works, VQA task is typically formulated as a multi-class classification problem. Given a dataset $\mathcal{D}_{\text{orig}} = \{ I_i, Q_i, a_i \}^N_i$ consisting of triplets of images $ I_i \in \mathcal{I} $, questions $ Q_i \in \mathcal{Q} $ and ground-truth answers $ a_i \in \mathcal{A}$, VQA model learns a multimodal mapping: $\mathcal{I} \times \mathcal{Q} \rightarrow [0, 1]^{|\mathcal{A}|}$, which produces an answer distribution given an image-question (VQ) pair.

To reduce the language biases, a surge of data augmentation (DA) methods have been proposed for VQA. Specifically, given the original training set $\mathcal{D}_{\text{orig}}$, DA methods generate an augmented training set $\mathcal{D}_{\text{aug}}$ automatically. Then, they can train any VQA architectures with both two training sets ($\mathcal{D}_{\text{orig}} \cup \mathcal{D}_{\text{aug}}$).

In this section, we first compare proposed KDDAug with existing DA methods in Sec.~\ref{sec:3.1}. Then, we introduce details of KDDAug, including image-question pair composition in Sec.~\ref{sec:3.2}, and KD-based answer assignment in Sec.~\ref{sec:3.3}.

\subsection{KDDAug vs. Existing DA Pipelines} \label{sec:3.1}

\noindent\textbf{Synthetic-based Methods.} For each human-labeled sample $(I_i, Q_i, a_i) \in \mathcal{D}_{\text{orig}}$, the synthetic-based methods (\eg, CSS~\cite{chen2020counterfactual}) always synthesize one corresponding augmented sample by either editing the image $I_i$ or question $Q_i$, denoted as $\hat{I}_i$ or $\hat{Q}_i$. Then, original image $I_i$ and its synthesized question $\hat{Q}_i$ compose a new VQ pair $(I_i, \hat{Q}_i)$ (similar for VQ pair $(\hat{I}_i, Q_i)$ and $(\hat{I}_i, \hat{Q}_i)$). Lastly, different answer assignment mechanisms are designed to generate pseudo answers for these augmented VQ pairs, and these samples constitute the augmented set $\mathcal{D}_{\text{aug}}$. As discussed in Sec.~\ref{sec:1}, these new synthetic VQ pairs are unnatural and error-prone.

\noindent\textbf{SimpleAug~\cite{kil2021discovering}.} Unlike synthetic-based methods, SimpleAug tries to compose new VQ pairs by randomly sampling an image $I_i \in \mathcal{I}$ and other possible question $Q_j \in \mathcal{Q}$, \ie, $(I_i, Q_j)$. This simple strategy can make sure both image $I_i$ and question $Q_j$ are always pristine. Not surprisingly, there is no free lunch --- this arbitrary composition strategy significantly increases the difficulty of pseudo answers assignment. To this end, SimpleAug proposes a set of heuristic rules for only four types of questions (\Qyesno, \Qnumber, \Qwhat, and \Qcolor), which limits its diversity and generalization ability.

\noindent\textbf{Proposed KDDAug.} To solve all the weaknesses in existing DA methods (\ie, both synthetic-based methods and SimpleAug), we take two steps to generate the augmented set $\mathcal{D}_{\text{aug}}$: 1) We randomly compose image $I_i \in \mathcal{I}$ and all reasonable questions $Q_j \in \mathcal{Q}$ without limiting question types. 2) We utilize a knowledge distillation (KD) based answer assignment to automatically generate ``soft" pseudo answers for each VQ pair. Next, we detailed introduce these two steps.

\subsection{Image-Question Pair Composition} \label{sec:3.2}

\begin{figure}[t]
	\centering
	\setlength{\abovecaptionskip}{-0.5em}
	\setlength{\belowcaptionskip}{0em}	\includegraphics[width=0.98\linewidth]{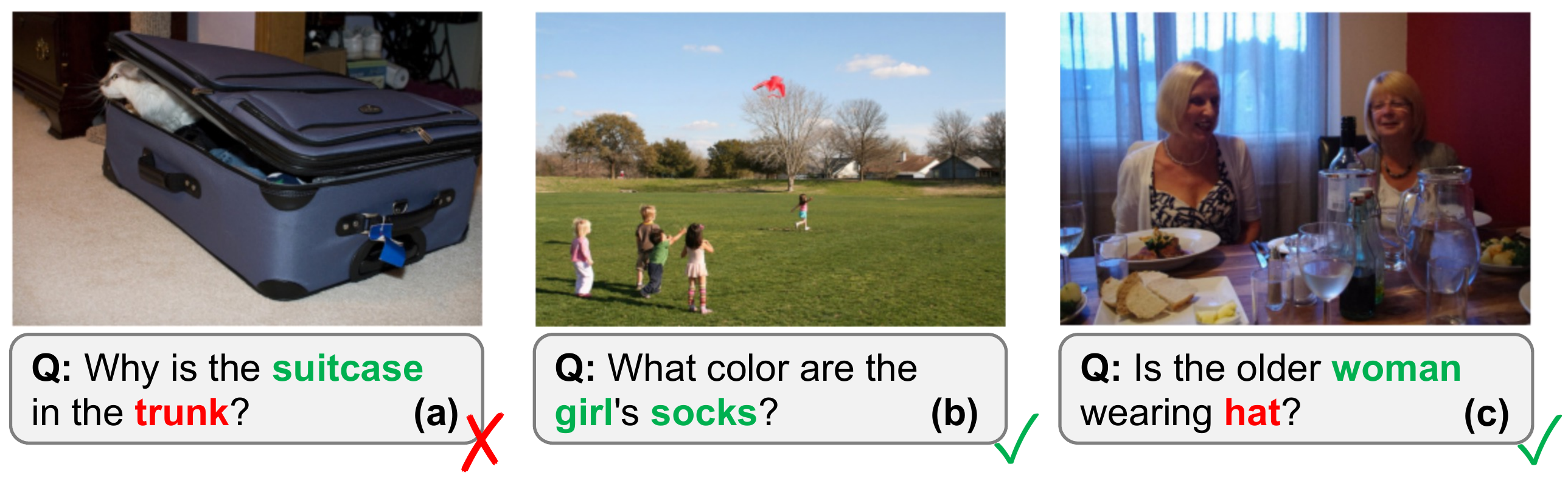}
	\caption{Example of three randomly composed VQ pairs. (a) \textbf{Unreasonable pair}: The question contains noun ``\texttt{trunk}" which are not in the image. (b) \textbf{Reasonable pair}: All nouns in the question (``\texttt{girl}" and ``\texttt{sock}") are in the image. (c) \textbf{Reasonable pair}: Although the question contains ``\texttt{hat}" which are not in the image, it is still reasonable.}
	\label{fig:vq_pair_examples}
\end{figure}

To extremely increase the diversity of the new augmented training set $\mathcal{D}_{\text{aug}}$, we relax the requirements for \emph{reasonable} VQ pairs by only considering the object categories in the images and nouns in the questions. By ``reasonable", we mean that: 1) The question is suitable for the image content. 2) There are some ground-truth answers for this VQ pair. For example, as shown in Fig.~\ref{fig:vq_pair_examples}(a), the question ``\texttt{Why is the suitcase in the trunk?}" is not a reasonable question for the image, because ``\texttt{truck}" does not appear in the image. Therefore, we treat question $Q_j$ as a reasonable question for image $I_i$ as long as all the meaningful nouns in $Q_j$ appear in $I_i$. For example in Fig.~\ref{fig:vq_pair_examples}(b), questions only containing ``\texttt{girl}" and ``\texttt{sock}" are all reasonable questions for this image\footnote{Some questions containing extra nouns may also be reasonable questions, especially for \Qyesno~questions (\cf~example in Fig.~\ref{fig:vq_pair_examples}(c)). However, almost all VQ pairs that meet this more strict requirement are always reasonable.}.

Thus, similar to other DA methods~\cite{kil2021discovering,chen2020counterfactual,chen2021counterfactual}, we first extract these meaningful nouns from all questions $\mathcal{Q}$. We utilize the spaCy POS tagger~\cite{honnibal2017spacy} to extract all nouns and unify their singular and plural forms\footnote{We tried to use WordNet to map between nouns' synsets, lemmas or hypernyms (\eg, ``dog" and ``animal"). But empirically, the VQA performance is quite similar.}. We ignore the nouns such as ``picture" or ``photo". We remove all the questions without any meaningful nouns (the proportion is small, \eg, $\approx$ 9\% in VQA-CP v2). For all images $\mathcal{I}$, we leverage an off-the-shelf object detector to detect all proposals in each image, and predict their object categories. Lastly, we compose all possible reasonable VQ pairs by traversing all the questions and images in the original training set.

Since the number of \Qyesno~questions is quite large (\eg, $\approx$ 42\% in VQA-CP v2), to prevent creating too many \Qyesno~samples, we group all ``\texttt{Yes/No}" questions with the same set of nouns into one group, and randomly select three questions per group for new sample compositions.

\noindent\textbf{CLIP-based Filtering.} One potential weakness for our VQ pair composition strategy is the excessive training samples, which may increase the training times. To achieve a good trade-off between efficiency and effectiveness, we can utilize a pretrained visual-language model CLIP~\cite{radford2021learning} to filter out less-efficient augmented samples. By ``less-efficient", we mean that the improvements provided by these training samples are marginal. This filtering design is based on the observation that people tend to ask questions about salient objects in the image, \ie, the nouns mentioned in questions should appear prominently in the image. Specifically, we firstly use the template ``\texttt{a photo of $<$NOUN$>$}" to generate prompts for each meaningful noun in the question and utilize CLIP to calculate the similarity score between the image and all corresponding prompts. Then, we use the average similarity score over all meaningful nouns to get the relevance score between the question and the image. We sort all composed VQ pairs according to the relevance score and only the $\alpha$\% samples with the top highest relevance scores are reserved (See Table~\ref{tab:CLIP2} for more details about influence of different $\alpha$). 

\noindent\textbf{Advantages.} Compared with existing VQ pair composition strategies, ours has several advantages: 1) Our definition for ``reasonable" simplifies the composition step and further improves the diversity of the VQ pairs. 2) Our strategy gets rid of human annotations, which means it can be easily extended to other datasets.

\begin{figure}[t]
	\centering
	\setlength{\abovecaptionskip}{0em}
	\setlength{\belowcaptionskip}{-1.0em}
	\includegraphics[width=\linewidth]{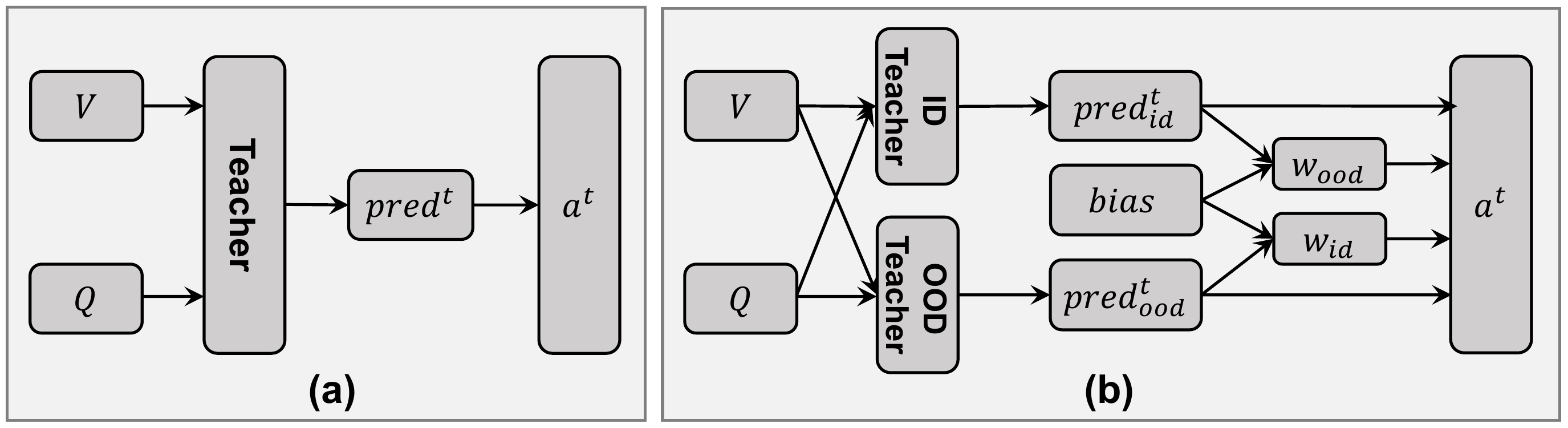}
	\caption{Pipelines of single-teacher KD (a) and multi-teacher KD (b) answer assignment.}
	\label{fig:answer_assignment}
\end{figure}

\subsection{KD-based Answer Assignment} \label{sec:3.3}

Given the original training set $\mathcal{D}_{\text{orig}}$ and all new composed reasonable VQ pairs (\{$I_i, Q_j$\}), we use a KD-based answer assignment to generate pseudo answers.

\noindent\textbf{Single-Teacher KD for Answer Assignment.}
Inspired by existing KD work for pseudo labeling~\cite{radosavovic2018data,niu2021introspective}, we begin to shift our gaze from manual rules to KD. A straightforward KD-based strategy is training a \emph{teacher} VQA model with the original training set $\mathcal{D}_{\text{orig}}$, and then using the pre-trained teacher model to predict answer distributions $pred^t$ for each composed VQ pair. And the $pred^t$ is treated as the pseudo answer for this VQ pair (\cf~Fig.~\ref{fig:answer_assignment}(a)). Obviously, the quality of assigned answers is determined by the performance of the teacher model, and the biases learned by the teacher model are also included in $pred^t$.

\noindent\textbf{Multi-Teacher KD for Answer Assignment.} To generate more accurate and robust pseudo answers, an extension is merging knowledge from multiple teachers. Given $N$ pre-trained teacher models, and each trained model can predict an answer distribution $\{pred^t_i\}$, and the pseudo answer is:
\begin{equation}
    a^{t} = \textstyle{\sum}^N_{i=1} w_i * pred^t_i, \;\; w.r.t \;\; \sum^N_{i=1} w_i=1,
    \label{eq:weighted_sum}
\end{equation}
where $w_i$ denotes the weight of $i$-th teacher model.

To make pseudo ground-truth answers more informative and robust to both ID and OOD benchmarks, in KDDAug, we adopt two expert teacher models: \emph{ID teacher} and \emph{OOD teacher}. Among them, ID and OOD teachers can effectively extract ID and OOD knowledge, respectively. Given the predicted answer distributions of ID teacher and OOD teacher (denoted as $pred^t_{id}$ and $pred^t_{ood}$), we then need to calculate the weights for the two teachers. Following Niu~\etal~\cite{niu2021introspective}, to obtain unbiased labels, we also assign a smaller weight to a more biased teacher. Since we lack any human-annotated ground-truth answers for these composed VQ pairs, we directly measure the bias degree of each teacher by calculating the cross-entropy (\texttt{XE}) loss between question-type bias (\emph{bias}) and their predictions:
\begin{equation}
    c_{id} = \frac{1}{\texttt{XE}(bias, pred^t_{id})}, \;\; c_{ood} = \frac{1}{\texttt{XE}(bias, pred^t_{ood})}, 
    \label{eq_confidence_calculation}
\end{equation}
where $bias$ is the statistical answer distribution of each question type, which is calculated from the original training set $\mathcal{D}_{\text{orig}}$. Obviously, if the prediction is more closer to $bias$, the teacher is more like to be biased. Then, we obtain:
\begin{equation}
    w_{id} = c_{ood}/ (c_{id} + c_{ood}), \;\; w_{ood} = c_{id} / (c_{id} + c_{ood}).
    \label{eq:weight_calculation}
\end{equation}
Lastly, we obtain pseudo answers $a^t$ by Eq.~\eqref{eq:weighted_sum} for each VQ pair (\cf~Fig.~\ref{fig:answer_assignment}(b)).

\noindent\textbf{Benefiting from Rule-based Initial Answers.} Based on Eq.~\eqref{eq_confidence_calculation}, our answer assignment can be easily extended to further benefit from high-quality rule-based initial answers (\ie, replacing \emph{bias} with high-quality initial answers). We denote these initial pseudo ground-truth answers as $a^{init}$. Following SimpleAug~\cite{kil2021discovering}, we consider three types of questions with a single noun:

1) \textbf{\Qcolor~questions.} For each paired image, the detector may output some color attributes. We assign the color of the noun in the question as $a^{init}$. 

2) \textbf{\Qnumber~questions.} For each paired image, we assign the count of detected objects which are same as the noun in the question as $a^{init}$. 

3) \textbf{\Qwhat~questions.} For each original sample $(I_i, Q_i, a_i)$, and new paired image $I_j$ for $Q_i$, if $a_i$ is in $I_j$'s object labels, we assign $a_i$ as the initial answer for VQ pair $(I_j, Q_i)$. For example, if original sample is ``\texttt{What is near the fork? Knife.}" If paired image contains ``\texttt{knife}", we assign ``\texttt{knife}" as $a^{init}$. 

We refer the readers to SimpleAug paper~\cite{kil2021discovering} for more details. After obtaining $a^{init}$, we can replace \emph{bias} to $a^{init}$ in these question types. Then, Eq.~\eqref{eq_confidence_calculation} becomes:
\begin{equation}
    c_{id} = \frac{1}{\texttt{XE}(a^{init}, pred^t_{id})}, \;\; c_{ood} = \frac{1}{\texttt{XE}(a^{init}, pred^t_{ood})}.
    \label{eq_confidence_calculation_v2}
\end{equation}
Considering that the initial answer $a^{init}$ is more accurate than the $pred^t_{id}$ \eg, ID teacher's ID performance is only 63.01\%~(\cf~Table~\ref{tab:teacher_effectiveness}), we follow~\cite{niu2021introspective} and use $a^{init}$ as ID knowledge, \ie, $a^{t} = w_{id} * a^{init} + w_{ood} * pred^t_{ood}$.

\noindent\textbf{Advantages.} Compared with the existing answer assignment mechanism, our solution gets rid of heuristic rules and human annotations. Meanwhile, it can be easily extended to generate better answers with more advanced teacher models. Besides, it is more general, which can theoretically be applied to any VQ pair.

\noindent\textbf{Why KDDAug can work?} KDDAug improves performance from two aspects: 1) It composes new samples to increase the diversity of the training set, which implicitly mitigates the biases with more balanced data. 2) It assigns more robust and informative answers for \emph{new} samples.

\section{Experiments}

\subsection{Experimental Settings and Implementation Details}\label{sec:4.1}

\textbf{Evaluation Datasets.} We evaluated the proposed KDDAug on two datasets: the ID benchmark \textbf{VQA v2}~\cite{goyal2017making} and OOD benchmark \textbf{VQA-CP v2}~\cite{agrawal2018don}. For model accuracies, we followed the standard VQA evaluation metric~\cite{antol2015vqa}. Meanwhile, we followed~\cite{niu2021counterfactual} and used Harmonic Mean~(\textbf{HM}) to evaluate the trade-off between ID and OOD evaluations. More details are left in the appendix.

\noindent\textbf{VQA Models.} Since KDDAug is an architecture-agnostic DA method, we evaluated the effectiveness of KDDAug on multiple different VQA models: UpDn~\cite{anderson2018bottom}, LMH~\cite{clark2019don}, RUBi~\cite{cadene2019rubi} and CSS~\cite{chen2020counterfactual,chen2021counterfactual}. Specifically, UpDn is a simple but effective VQA model, which always serves as a backbone for other advanced VQA models. LMH, RUBi, and CSS are SOTA ensemble-based VQA models for debiasing. For each specific VQA baseline, we followed their respective configurations (\eg, hyperparameter settings) and re-implemented them using the official codes.

\noindent\textbf{ID \& OOD Teachers.} The ID and OOD teachers were from a same LMH-CSS~\cite{chen2020counterfactual} model with different architectures~\cite{niu2021counterfactual,niu2021introspective}. Since LMH-CSS is an ensemble-based debiasing model, we took the whole ensemble model (VQA w/ bias-only model) as ID teacher, and the bare VQA model as OOD teacher (original for debiasing). Benefiting from the different architectures, they can extract ID and OOD knowledge, respectively. We used the official LMH-CSS codes to train ID and OOD teachers simultaneously \emph{on a same dataset}, \eg, for VQA-CP evaluation, both teachers were trained on the VQA-CP training set.

\noindent\textbf{Two Augmented Dataset Versions.} Due to the huge amount of all reasonable composed VQ pairs, and to keep fair comparisons with existing DA methods (especially SimpleAug), we constructed two versions of augmented sets: 1) $D^{\text{basic}}_{\text{aug}}$: It only contains the same four types of questions as SimpleAug. Meanwhile, it also contains a same set of extra training samples by paraphrasing\footnote{Paraphrasing is an supplementary DA tricks proposed by SimpleAug~\cite{kil2021discovering}. Specifically, for each original sample $(I_i, Q_i, a_i)$, if question $Q_j$ is similar to $Q_i$ (predicted by pre-trained BERT~\cite{devlin2019bert}), they construct a new augmented training sample $(I_i, Q_j, a_i)$. \label{footnote:paraphrasing}}. Thus, $D^{\text{basic}}_{\text{aug}}$ can clearly demonstrate the effectiveness of our proposed KD-based answer assignment strategy. 2) $D^{\text{extra}}_{\text{aug}}$: It contains all possible question types. Obviously, all VQ pairs from paraphrasing\footref{footnote:paraphrasing} are a subset of $D^{\text{extra}}_{\text{aug}}$. To decrease the number of augmented samples, we applied the CLIP-based filtering (\cf~Sec.~\ref{sec:3.2}) to keep top 10~\% samples. In the following experiments, we denote the model trained with $D^{\text{basic}}_{\text{aug}}$ and $D^{\text{extra}}_{\text{aug}}$ as \textbf{KDDAug} and \textbf{KDDAug$^+$}, respectively. Meanwhile, unless otherwise specified, we used the rule-based initial answers for $D^{\text{basic}}_{\text{aug}}$.

\noindent\textbf{Training Details \& KDDAug Settings.} Details are left in the appendix.

%%%%%%%%%%%%% Architecture Agnostic %%%%%%%%%%%%%%%%
\begin{table*}[t]
	\setlength{\abovecaptionskip}{-0.5em}
	\setlength{\belowcaptionskip}{-0.5em}
	\begin{center}
	    \scalebox{0.85}{
	        \begin{tabular}{| l | l | c c c c | c c c c| c |}
			\hline
			\multirow{2}{*}{\textbf{Base}} & \multirow{2}{*}{\textbf{Models}} & \multicolumn{4}{c|}{VQA-CP v2 test} & \multicolumn{4}{c|}{VQA v2 val} &  \multirow{2}{*}{\textbf{HM}} \\
			& & \textbf{All} & Y/N & Num & Other & \textbf{All} & Y/N & Num & Other &  \\
		    \hline
		        \multirow{3}{*}{UpDn~\cite{anderson2018bottom}} & Baseline & 39.74 & 42.27 & 11.93 & 46.05 & 63.48 & 81.18 & 42.14 & 55.66 &  48.88 \\
		        & Baseline$^*$ & 39.85 & 42.66 & 12.18 & 45.98 & \textbf{63.30} & \textbf{81.06} & \textbf{42.46} & \textbf{55.32} & 48.91 \\
		        & KDDAug & \textbf{60.24}$_{\textcolor{mygreen}{\textbf{+20.39}}}$ & \textbf{86.13} & \textbf{55.08} & \textbf{48.08} & 62.86$_{\textcolor{red}{\textbf{-0.44}}}$ & 80.55 & 41.05 & 55.18 & \textbf{61.52}$_{\textcolor{mygreen}{\textbf{+12.61}}}$ \\
            \hline
                \multirow{3}{*}{LMH~\cite{clark2019don}} & Baseline & 52.05 & --- & --- & --- & --- & --- & --- & --- & --- \\
		        & Baseline$^*$ & 53.87 & 73.31 & 44.23 & 46.33 & 61.28 & 76.58 & \textbf{55.11} & 40.69 & 57.24 \\
		        & KDDAug & \textbf{59.54}$_{\textcolor{mygreen}{\textbf{+5.67}}}$ & \textbf{86.09} & \textbf{54.84} & \textbf{46.92} & 62.09$_{\textcolor{mygreen}{\textbf{+0.81}}}$ & \textbf{79.26} & 40.11 & \textbf{54.85} & \textbf{60.79}$_{\textcolor{mygreen}{\textbf{+3.55}}}$ \\
            \hline
                \multirow{3}{*}{RUBi~\cite{cadene2019rubi}} & Baseline & 44.23 & --- & --- & --- & --- & --- & --- & --- & --- \\
		        & Baseline$^*$ & 46.84 & 70.05 & 44.29 & 11.85 & 52.83 & 54.74 & \textbf{41.56} & \textbf{54.38} & 49.66 \\
		        & KDDAug & \textbf{59.25}$_{\textcolor{mygreen}{\textbf{+12.41}}}$ & \textbf{84.16} & \textbf{54.12} & \textbf{47.61} & \textbf{60.25}$_{\textcolor{mygreen}{\textbf{+7.42}}}$ & \textbf{74.97} & 40.29 & 54.35 & \textbf{59.75}$_{\textcolor{mygreen}{\textbf{+10.09}}}$ \\
		    \hline
                \multirow{3}{*}{CSS$^+$~\cite{chen2021counterfactual}} & Baseline & 59.54 & 83.37 & 52.57 & 48.97 & 59.96 & 73.69 & 40.18 & 54.77 & 59.75 \\
		        & Baseline$^*$ & 59.19 & 83.54 & 51.29 & \textbf{48.59} & 58.91 & 71.02 & 39.76 & \textbf{54.77} & 59.05 \\
		        & KDDAug & \textbf{61.14}$_{\textcolor{mygreen}{\textbf{+1.95}}}$ & \textbf{88.31} & \textbf{56.10} & 48.28 & \textbf{62.17}$_{\textcolor{mygreen}{\textbf{+3.26}}}$ & \textbf{79.50} & \textbf{40.57} & 54.71 & \textbf{61.65}$_{\textcolor{mygreen}{\textbf{+2.60}}}$ \\
			\hline
    		\end{tabular}
	    }
	\end{center}
	\caption[]{Accuracies (\%) on VQA-CP v2 and VQA v2 of different VQA architectures. $^*$ indicates the results from our reimplementation using official codes.}
	\label{tab:architecture_agnostic}
\end{table*}

\subsection{Architecture Agnostic}\label{sec:4-2}

\noindent\textbf{Settings.} Since KDDAug is a model-agnostic data augmentation method, it can seamlessly incorporated into any VQA architectures. To validate the generalization of KDDAug, we applied it to multiple different VQA models: UpDn~\cite{anderson2018bottom}, LMH~\cite{clark2019don}, RUBi~\cite{cadene2019rubi} and CSS$^+$~\cite{chen2021counterfactual}. All the results are shown in Table~\ref{tab:architecture_agnostic}.

\noindent\textbf{Results.} Compared to these baseline models, KDDAug can consistently improve the performance for all architectures, and push all models' performance to the state-of-the-art level. Particularly, the improvements are most signiﬁcant in the baseline UpDn model (\eg, 12.61\% absolute performance gains on HM). Furthermore, when KDDAug is applied to another DA-based model CSS$^+$, KDDAug can still improve the performance on both OOD and ID benchmarks, and achieve the best performance (\eg, 61.65\% on HM).

%%%%%%%%%%%%% STOA on VQA-CP v2 %%%%%%%%%%%%%%%%
\begin{table*}[t]
	\setlength{\abovecaptionskip}{-0.5em}
	\setlength{\belowcaptionskip}{-1.0em}
	\begin{center}
	    \scalebox{0.90}{
    		\begin{tabular}{| l | c | c c c c | c c c c| c |}
			\hline
    			\multirow{2}{*}{\textbf{Models}}  & \multirow{2}{*}{~\textbf{DA}~} & \multicolumn{4}{c|}{VQA-CP v2 test} & \multicolumn{4}{c|}{VQA v2 val} &  \multirow{2}{*}{\textbf{HM}} \\
	    		& & \textbf{All} & Y/N & Num & Other & \textbf{All} & Y/N & Num & Other &  \\
		    \hline
		    	UpDn~\cite{anderson2018bottom}$_{\texttt{CVPR'18}}$ & & 39.74 & 42.27 & 11.93 & 46.05 & 63.48 & 81.18 & 42.14 & 55.66 & 48.88\\
    			
    			~~+AReg~\cite{ramakrishnan2018overcoming}$_{\texttt{NeurIPS'18}}$ &  & 41.17 & 65.49 & 15.48 & 35.48 & 62.75 & 79.84 & 42.35 & 55.16 & 49.72 \\
    			
    			~~+MuRel~\cite{cadene2019murel}$_{\texttt{CVPR'19}}$ & & 39.54 & 42.85 & 13.17 & 45.04 & --- & --- & --- & --- &  --- \\
			    
			    ~~+GRL~\cite{grand2019adversarial}$_{\texttt{ACL'19}}$ & & 42.33 & 59.74 & 14.78 & 40.76 & 51.92 & --- & --- & --- & 46.64 \\
			    
			    ~~+CF-VQA~\cite{niu2021counterfactual}$_{\texttt{CVPR'21}}$ & & 53.55 & {91.15} & 13.03 & 44.97 & 63.54 & 82.51 & 43.96 & 54.30 & 58.12 \\

			    ~~+GGE-DQ~\cite{han2021greedy}$_{\texttt{ICCV'21}}$ & & 57.32 & 87.04 & 27.75 & 49.59 & 59.11 & 73.27 & 39.99 & 54.39 & 58.20 \\
			    
			    ~~+D-VQA~\cite{wen2021debiased}$_{\texttt{NeurIPS'21}}$ & & 61.91 & 88.93 & 52.32 & 50.39 & 64.96 & 82.18 & 44.05 & 57.54 & 63.40 \\
			 \cline{3-11}
			 	~~+CVL~\cite{abbasnejad2020counterfactual}$_{\texttt{CVPR'20}}$ & \boldcheckmark & 42.12 & 45.72 & 12.45 & 48.34 &  --- & --- & ---  & --- & --- \\
			 
			 	~~+Unshuffling~\cite{teney2020unshuffling}$_{\texttt{ICCV'21}}$ & \boldcheckmark & 42.39 & 47.72 & 14.43 & 47.24 & 61.08 & 78.32 & 42.16 & 52.81 & 50.05 \\
			 
			    ~~+CSS~\cite{chen2020counterfactual}$_{\texttt{CVPR'20}}$ & \boldcheckmark & 41.16 & 43.96 & 12.78 & 47.48 & --- & --- & --- & --- & --- \\
                
                ~~+CSS$^+$~\cite{chen2021counterfactual}$_{\texttt{arXiv'21}}$ & \boldcheckmark & 40.84 & 43.09 & 12.74 & 47.37 & --- & --- & --- & --- & --- \\
                
                ~~+RandImg~\cite{teney2020value}$_{\texttt{NeurIPS'20}}$ & \boldcheckmark & 55.37 & 83.89 & 41.60 & 44.20 & 57.24 & 76.53 & 33.87 & 48.57 & 56.29 \\
                
    			~~+SSL~\cite{zhu2020overcoming}$_{\texttt{IJCAI'20}}$ & \boldcheckmark & 57.59 & 86.53 & 29.87 & 50.03 & 63.73 &  --- & --- & --- & 60.50 \\
                
                ~~+MUTANT$^\dagger$~\cite{gokhale2020mutant}$_{\texttt{EMNLP'20}}$ & \boldcheckmark & 50.16 & 61.45 & 35.87 & 50.14 & --- & --- & --- & --- & --- \\
                
                ~~+SimpleAug~\cite{kil2021discovering}$_{\texttt{EMNLP'21}}$ & \boldcheckmark & 52.65 & 66.40 & 43.43 & 47.98 & \textbf{64.34} & \textbf{81.97} & \textbf{43.91} & \textbf{56.35} & 57.91 \\
            
                ~~+\textbf{KDDAug} & \boldcheckmark & \textbf{60.24} & \textbf{86.13} & \textbf{55.08} & \textbf{48.08} & 62.86 & 80.55 & 41.05 & 55.18 & \textbf{61.52} \\
            \hline
            	LMH$^*$~\cite{clark2019don}$_{\texttt{EMNLP'19}}$ & & 53.87 & 73.31 & 44.23 & 46.33 & 61.28 & 76.58 & 55.11 & 40.69 & 57.24 \\

                ~~+IntroD~\cite{niu2021introspective}$_{\texttt{NeurIPS'21}}$ & & 51.31 & 71.39 & 27.13 & 47.41 & 62.05 & 77.65 & 40.25 & 55.97 & 56.17 \\		
                
            	~~+CSS+CL~\cite{liang2020learning}$_{\texttt{EMNLP'20}}$ & & 59.18 & 86.99 & 49.89 & 47.16 & 57.29 & 67.27 & 38.40 & 54.71 & 58.22 \\
                
                ~~+CSS+IntroD~\cite{niu2021introspective}$_{\texttt{NeurIPS'21}}$ & & 60.17 & 89.17 & 46.91 & 48.62 & 62.57 & 78.57 & 41.42 & 56.00 & 61.35 \\
            \cline{3-11}
                ~~+CSS~\cite{chen2020counterfactual}$_{\texttt{CVPR'20}}$ & \boldcheckmark & 58.95 & 84.37 & 49.42 & 48.21 & 59.91 & 73.25 & 39.77 & 55.11 & 59.43 \\
			
			    ~~+CSS$^+$~\cite{chen2021counterfactual}$_{\texttt{arXiv'21}}$ & \boldcheckmark & 59.54 & 83.37  & 52.57  & 48.97  & 59.96 & 73.69  & 40.18  & 54.77  & 59.75 \\
			
			    ~~+SimpleAug~\cite{kil2021discovering}$_{\texttt{EMNLP'21}}$ & \boldcheckmark & 53.70 & 74.79 & 34.32 & 47.97 & \textbf{62.63} & 79.31 & \textbf{41.71} & \textbf{55.48} & 57.82 \\
			
			    ~~+ECD~\cite{kolling2022efficient}$_{\texttt{WACV'22}}$ & \boldcheckmark & 59.92 & 83.23 & 52.29 & \textbf{49.71} & 57.38 & 69.06 & 35.74 & 54.25 & 58.62 \\
			
			    ~~+\textbf{KDDAug} & \boldcheckmark & 59.54 & 86.09 & 54.84 & 46.92 & 62.09 & 79.26 & 40.11 & 54.85 & 60.79 \\
			
			    ~~+\textbf{CSS$^+$}+\textbf{KDDAug} & \boldcheckmark & \textbf{61.14} & \textbf{88.31} & \textbf{56.10} & 48.28 & 62.17 & \textbf{79.50} & 40.57 & 54.71 & \textbf{61.65} \\ 
            \hline
    		\end{tabular}
	    }
	\end{center}
	\caption[]{Accuracies (\%) on VQA-CP v2 and VQA v2 of SOTA models. ``\textbf{DA}" denotes the data augmentation methods. $^*$ indicates the results from our reimplementation. ``MUTANT$^\dagger$" denotes MUTANT~\cite{gokhale2020mutant} only trained with \texttt{XE} loss.} 
	\label{tab:SOTA_v2}
\end{table*}

\subsection{Comparisons with State-of-the-Arts}

\noindent{\textbf{Settings.}} We incorporated the KDDAug into model UpDn~\cite{anderson2018bottom}, LMH~\cite{clark2019don} and CSS$^+$~\cite{chen2021counterfactual}, and compared them with the SOTA VQA models both on VQA-CP v2 and VQA v2. According to the model framework design, we group them into: 1) \emph{Non-DA Methods}: UpDn~\cite{anderson2018bottom}, AReg~\cite{ramakrishnan2018overcoming}, MuRel~\cite{cadene2019murel}, GRL~\cite{grand2019adversarial}, CF-VQA~\cite{niu2021counterfactual}, GGE-DQ~\cite{han2021greedy}, D-VQA~\cite{wen2021debiased}, IntroD~\cite{niu2021introspective}, CSS+CL~\cite{liang2020learning}, and LMH~\cite{clark2019don}. 2) \emph{DA Methods}: CVL~\cite{abbasnejad2020counterfactual}, Unshuffling~\cite{teney2020unshuffling}, CSS~\cite{chen2020counterfactual}, CSS$^+$~\cite{chen2021counterfactual}, RandImg~\cite{teney2020value}, SSL~\cite{zhu2020overcoming}, MUTANT~\cite{gokhale2020mutant}, SimpleAug~\cite{kil2021discovering}, and ECD~\cite{kolling2022efficient}. All results are reported in Table~\ref{tab:SOTA_v2}.

\noindent{\textbf{Results.}} Compared with all existing DA methods, KDDAug achieves the best OOD and trade-off performance on two datasets. For UpDn backbone, KDDAug improves the OOD performance of UpDn with a 20\% absolute performance gain (60.24\% vs. 39.74\%) and improves accuracies on all different question categories. For LMH backbone, KDDAug boosts the performance on both ID and OOD benchmarks. Compared with other non-DA methods, KDDAug still outperforms most of them. It is worth noting that our KDDAug can also be incorporated into these advanced non-DA models to further boost their performance.

\subsection{Ablation Studies}\label{sec:4.4}

We validate the effectiveness of each component of KDDAug by answering the following questions: \textbf{Q1}: Does KDDAug assign more robust answers than existing methods? \textbf{Q2}: Does KDDAug mainly rely on the rule-based initial answers? \textbf{Q3}: Does KDDAug only benefit from much more training samples? \textbf{Q4}: Does the multi-teacher design help to improve pseudo ground-truth answers quality? \textbf{Q5}: Is the diversity of question types important for the data augmentation model?

\textbf{KDDAug vs. SimpleAug~\cite{kil2021discovering} (Q1).} To answer \textbf{Q1}, we compared the answers assigned by KDDAug and SimpleAug. Due to different composition strategies for \Qyesno~questions, we firstly removed the \Qyesno~questions from $D^{\text{basic}}_{\text{aug}}$. We use a pretrained CLIP~\cite{radford2021learning} (denoted as CLIP$_{\text{rank}}$ and more details are left in the appendix.) to rank the quality of all SimpleAug assigned answers. For more comprehensive comparisons, we divided the augmented samples into three subsets according to the ranks: 1) All augmented samples (100\%), 2) Top-50\% samples~($\uparrow$~50\%), and 3) Bottom-50\% samples~($\downarrow$~50\%). We compared KDDAug and SimpleAug on these three subsets, and results are in Table~\ref{tab:SimpleAug}.

\textit{\textbf{Results for Q1.}} From Table~\ref{tab:SimpleAug}, we can observe: 1) The performance of SimpleAug on three subsets varies significantly, \eg, bottom-50\% samples lead to a huge drop in HM (-0.87\%). 2) In contrast, KDDAug on different subsets achieves similar decent performance. 3) With the same augmented image-question pairs, KDDAug consistently outperforms the corresponding SimpleAug by a significant margin~(over 8\% on HM), which proves the robustness of our answers.

%%%%%%%%%%%%% KDDAug vs. SimpleAug %%%%%%%%%%%%%%%%
\begin{table*}[t]
	\setlength{\abovecaptionskip}{-0.5em}
	\setlength{\belowcaptionskip}{-0.5em}
	\begin{center}
        \scalebox{0.95}{
        	\begin{tabular}{|l | c c c c | c c c c| c |}
			\hline
			\multirow{2}{*}{Models} & \multicolumn{4}{c|}{VQA-CP v2 test} & \multicolumn{4}{c|}{VQA v2 val} & \multirow{2}{*}{\textbf{HM}} \\
			& \textbf{All} & Yes/No & Num & Other & \textbf{All} & Yes/No & Num & Other &  \\
		    \hline
		    UpDn~\cite{anderson2018bottom} & 39.74 & 42.27 & 11.93 & 46.05 & 63.48 & 81.18 & 42.14 & 55.66 & 48.88 \\
		    ~~+SimpleAug~\cite{kil2021discovering} & 52.65 & 66.40 & 43.43 & 47.98 & 64.34 & 81.97 & 43.91 & 56.35 & 57.91 \\
		    ~~+SimpleAug$^*$ & 48.56 & 58.83 & 35.41 & 46.78 & 60.67 & 75.65 & 40.45 & 54.65 & 54.57\\
		    ~~+SimpleAug$^-$ (100\%) & 45.38 & 45.59 & 37.63 & 47.40 & 62.62 & 80.49 & 40.68 & 54.85 & 52.62 \\
		    ~~+SimpleAug$^-$ ($\uparrow$~50\%) & 46.35 & 47.72 & 39.36 & 47.55 & 62.51 & 80.64 & 40.08 & 54.67 & 53.23 \\
		    ~~+SimpleAug$^-$ ($\downarrow$~50\%) & 45.06 & 44.84 & 38.44 & 46.99 & 62.49 & 80.38 & 40.24 & 54.78 & 52.36\\
			\hline\hline
            ~~+\textbf{KDDAug} & 60.24 & 86.13 & 55.08 & 48.08 & 62.86 & 80.55 & 41.05 & 55.18 & 61.52\\
            ~~+\textbf{KDDAug}$^-$ (100\%) & 59.96 & 84.95 & 54.98 & 48.23 & 62.72 & 80.07 & 40.90 & 55.30 & 61.31\\
            ~~+\textbf{KDDAug}$^-$ ($\uparrow$~50\%) & 59.94 & 84.78 & 54.70 & 48.36 & 62.67 & 79.86 & 41.06 & 55.32 & 61.27\\
            ~~+\textbf{KDDAug}$^-$ ($\downarrow$~50\%) & 59.97 & 85.11 & 55.13 & 48.13 & 62.60 & 80.13 & 40.78 & 55.04 & 61.26\\
			\hline
    		\end{tabular}
        }
	\end{center}
	\caption[]{Accuracies (\%) on different augmented subsets. $^*$ indicates our reimplementation. For fair comparisons, SimpleAug$^*$ didn't leverage human annotation and didn't remove examples that can be answered. $^-$ denotes without \Qyesno~questions.}
	\label{tab:SimpleAug}
\end{table*}

\begin{table*}[t]
	\setlength{\abovecaptionskip}{-0.5em}
	\setlength{\belowcaptionskip}{-1.0em}
	\begin{center}
	    \scalebox{0.95}{
    		\begin{tabular}{|l | c | c c c c c c c c c c c|}
                \hline
                    & Baseline & \multicolumn{11}{c|}{KDDAug} \\
                    & (UpDn) & 0\% & 10\% & 20\% & 30\% & 40\% & 50\% & 60\% & 70\% & 80\% & 90\% & 100\% \\
                \hline
                    VQA-CP v2 & 39.74 & 53.99 & 54.37 & 54.98 & 56.17 & 57.51 & 58.80 & 59.27 & 59.68 & 60.02 & 60.19 & 60.24 \\
                    VQA v2 & 63.48 & 63.26 & 63.24 & 63.20 & 63.20 & 63.19 & 63.10 & 63.08 & 63.04 & 63.01 & 62.83 & 62.86 \\
                \hline
                    HM & 48.88 & 58.26 & 58.47 & 58.80 & 59.48 & 60.22 & 60.87 & 61.12 & 61.31 & 61.48 & 61.48 & 61.52 \\
    			\hline
    		\end{tabular}
	    }
	\end{center}
	\caption[]{Accuracies (\%) on VQA-CP v2 test set and VQA v2 valset  of different $\delta$.} 
	\label{tab:CLIP1}
\end{table*}

\textbf{Influence of Rule-based Initial Answers (Q2).} To evaluate the quality of directly automatically assigned answers, we again used CLIP$_{\text{rank}}$ to divide the augmented samples (except \Qyesno~samples) into two parts. The augmented samples with top-$\delta$\% ranks use rule-based initial answers (\cf~Eq.~\eqref{eq_confidence_calculation_v2}) and left samples use question-type \emph{bias} (\cf~Eq.~\eqref{eq_confidence_calculation}). All results are shown in Table~\ref{tab:CLIP1}.

\textit{\textbf{Results for Q2.}} From the results, we can observe that KDDAug achieves consistent gains against the baseline (UpDn) on all proportions. The performance is best when all samples use rule-based initial samples. Even if all the samples' answers are assigned without any initial answers, KDDAug still gains significant improvement gains (58.26\% vs. 48.88\% on HM), and is better than SimpleAug.

\begin{table*}[t]
	\setlength{\abovecaptionskip}{-0.5em}
	\setlength{\belowcaptionskip}{-1.0em}
	\begin{center}
	    \scalebox{0.9}{
    		\begin{tabular}{|l | c c c c l| c c c c l|c|}
			\hline
			\multirow{2}{*}{Models ($\alpha \%$)} & \multicolumn{5}{c|}{VQA-CP v2 test} & \multicolumn{5}{c|}{VQA v2 val} & \multirow{2}{*}{\textbf{HM}} \\
			& \textbf{All} & Y/N & Num & Other & \#Samples & \textbf{All} & Y/N & Num & Other & \#Samples & \\ 
		    \hline
		    UpDn~\cite{anderson2018bottom} & 39.74 & 42.27 & 11.93 & 46.05 & 438K & 63.48 & 81.18 & 42.14 & 55.66 & 444K & 48.88 \\
            ~+KDDAug (100\%) & 60.24 & 86.13 & 55.08 & 48.08 & ~+4,088K & 62.86 & 80.55 & 41.05 & 55.18 & ~+2,279K & 61.52 \\
            ~+KDDAug (90\%) & 60.19 & 86.09 & 55.13 & 48.00 & ~+3,679K & 62.83 & 80.53 & 41.14 & 55.12 & ~+2,051K & 61.48\\
            ~+KDDAug (70\%) & 60.12 & 85.96 & 55.09 & 47.96 & ~+2,861K & 62.82 & 80.52 & 40.99 & 55.14 & ~+1,595K & 61.44 \\
            ~+KDDAug (50\%) & 60.13 & 86.18 & 54.81 & 47.94 & ~+2,044K & 62.71 & 80.41 & 40.98 & 55.00 & ~+1,139K & 61.39\\
            ~+KDDAug (30\%) & 59.92 & 86.06 & 54.74 & 47.64 & ~+1,226K & 62.51 & 80.35 & 40.56 & 54.76 & ~+684K & 61.19 \\
            ~+KDDAug (10\%) & 59.41 & 85.81 & 54.85 & 46.82 & ~+409K & 62.37 & 80.47 & 40.98 & 54.28 & ~+228K & 60.85 \\
			\hline
            ~+SimgpleAug~\cite{kil2021discovering} & 52.65 & 66.40 & 43.43 & 47.98 & ~+3,081K & 64.34 & 81.97 & 43.91 & 56.35 & ~--- & 57.91\\
            ~+SimgpleAug$^*$ & 48.56 & 58.83 & 35.41 & 46.78 & ~+4,702K & 60.67 & 75.65 & 40.45 & 54.65 & ~+2,358K & 54.57\\
			\hline
    		\end{tabular}
	    }
	\end{center}
	\caption[]{Accuracies (\%) of different $\alpha$ in CLIP-based filtering. SimpleAug$^*$ is the same as Table~\ref{tab:SimpleAug}. ``\#Samples" denotes the number of total training samples.}
	\label{tab:CLIP2}
\end{table*}

%%%%%%%%%%%%% Effectiveness of Teacher %%%%%%%%%%%%%%%%
\begin{table*}[t]
	\setlength{\abovecaptionskip}{-0.5em}
	\setlength{\belowcaptionskip}{-1.0em}
	\begin{center}
	    \scalebox{0.94}{
    		\begin{tabular}{| l | c | c c c c | c c c c| c |}
    			\hline
    			    \multirow{2}{*}{\textbf{Models}} & \multirow{2}{*}{\textbf{OOD W.}} & \multicolumn{4}{c|}{VQA-CP v2 test} & \multicolumn{4}{c|}{VQA v2 val} & \multirow{2}{*}{\textbf{HM}} \\
    			    & & \textbf{All} & Y/N & Num & Other & \textbf{All} & Y/N & Num & Other &  \\
    			\hline
    			    UpDn~\cite{anderson2018bottom} (baseline) & & 39.74 & 42.27 & 11.93 & 46.05 & 63.48 & 81.18 & 42.14 & 55.66 & 48.88 \\
    		        ID-Teacher & & 36.93 & 36.56 & 12.82 & 43.73 & 63.01 & 80.76 & 42.30 & 55.01 & 46.57\\
    		        OOD-Teacher & & 58.07 & 82.47 & 52.03 & 46.93 & 60.21 & 74.19 & 40.32 & 54.86 & 59.12 \\
    		    \hline
    		        Simple Avg. & 0.5 & 53.06 & 63.89 & 50.72 & 48.03 & \textbf{63.45} & \textbf{81.33} & \textbf{42.48} & \textbf{55.41} & 57.79 \\
                    ID-distill. & 0.0 & 43.10 & 42.33 & 29.34 & 47.28 & 62.90 & 81.10 & 41.05 & 54.84 & 51.15 \\
                    OOD-distill. & 1.0 & 58.40 & 82.27 & 53.16 & 47.33 & 61.50 & 77.08 & 41.62 & 54.92 & 59.91 \\
                    \textbf{KDDAug (Ours)} & dynamic & \textbf{60.24} & \textbf{86.13} & \textbf{55.08} & \textbf{48.08} & 62.86 & 80.55 & 41.05 & 55.18 & \textbf{61.52} \\
                \hline
    	    \end{tabular}
	    }
	\end{center}
	\caption[]{Effects of different teachers on pseudo answer assignment. ``OOD W." denotes $w_{ood}$, and ``dynamic" denotes $w_{ood}$ is dynamically calculated by our strategy.}
	\label{tab:teacher_effectiveness}
\end{table*}

\textbf{Influence of Number of Augmented Samples (Q3).} We set different $\alpha$ values in the CLIP-based filtering to control the number of augmented samples.

\textit{\textbf{Results for Q3.}} From the results in Table~\ref{tab:CLIP2}, we have several observations: 1) When more samples are used, the model performs better, which reflects the robustness of generated pseudo answers. 2) Even if a small number of samples are used (\eg, $\alpha\% = 10\%$), KDDAug still achieves decent performance, and is better than SimpleAug~(60.85\% vs. 57.91\%). 3) By adjusting the value of $\alpha$, we can easily achieve a trade-off between training efficiency and model performance.

\begin{figure}[t]
	\centering
	\setlength{\abovecaptionskip}{0em}
	\setlength{\belowcaptionskip}{-0.5em}
	\includegraphics[width=\linewidth]{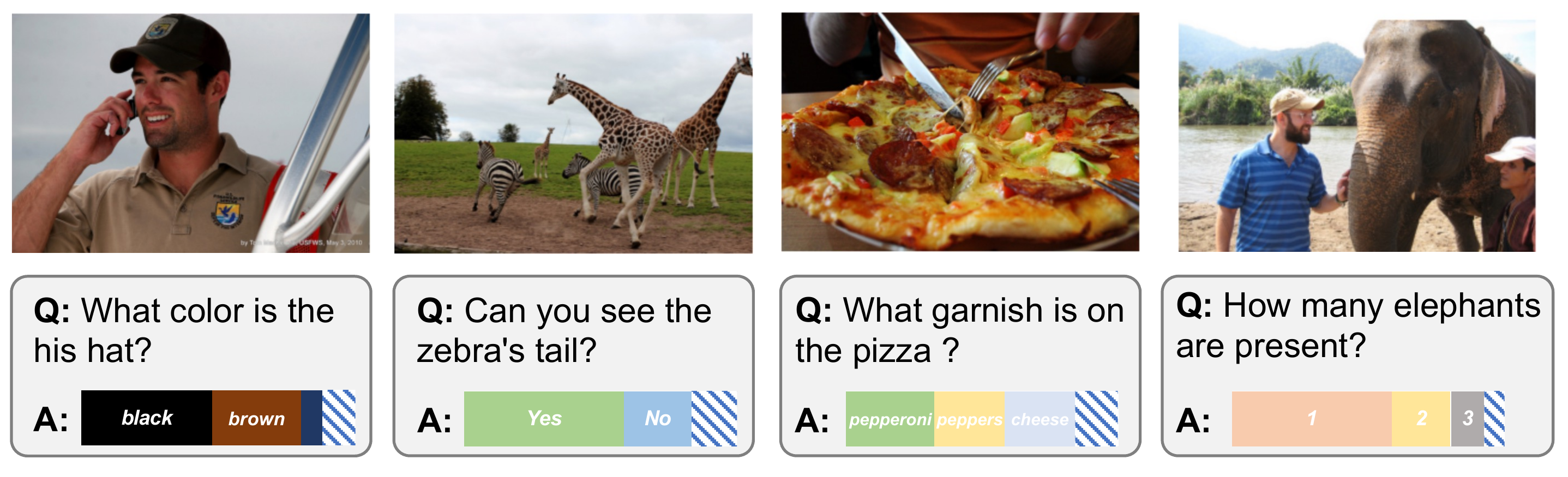}
	\caption{Visualization results of some augmented samples by our KDDAug.}
	\label{fig:visualization}
\end{figure}

\begin{table*}[t]
	\setlength{\abovecaptionskip}{-0.5em}
	\setlength{\belowcaptionskip}{-1.0em}
	\begin{center}
	    \scalebox{0.94}{
    		\begin{tabular}{|l | c | c c c c| c c c c |c|}
    			\hline
    			\multirow{2}{*}{\textbf{Models} ($\alpha \%$)} & \multirow{2}{*}{\textbf{Extra}} & \multicolumn{4}{c|}{VQA-CP v2 test} & \multicolumn{4}{c|}{VQA v2 val} & \multirow{2}{*}{\textbf{HM}} \\
    			& & \textbf{All} & Y/N & Num & Other & \textbf{All} & Y/N & Num & Other & \\ 
    		    \hline
    		    UpDn~\cite{anderson2018bottom} & & 39.74 & 42.27 & 11.93 & 46.05 & 63.48 & 81.18 & 42.14 & 55.66 & 48.88 \\
                \hline
                ~~+KDDAug$^\ddagger$~(100\%) & & 53.03 & \textbf{86.55} & \textbf{17.30} & 45.26 & 61.59 & 80.43 & 42.33 & 52.36 & 56.99\\
                ~~+KDDAug$^+$$^\ddagger$~(100\%) & \boldcheckmark & \textbf{53.76} & 86.41 & 17.10 & \textbf{46.70} & \textbf{62.58} & \textbf{80.51} & \textbf{42.43} & \textbf{54.28} & \textbf{57.84} \\
                \hline
                ~~+KDDAug$^\ddagger$~(50\%) & & 52.91 & 85.59 & 17.01 & 45.64 & 62.01 & 80.42 & 41.98 & 53.31 & 57.10\\
                ~~+KDDAug$^+$$^\ddagger$~(50\%) & \boldcheckmark & \textbf{53.76} & \textbf{86.57} & \textbf{17.86} & \textbf{46.42} & \textbf{62.64} & \textbf{80.35} & \textbf{42.25} & \textbf{54.56} & \textbf{57.86} \\
                \hline
                ~~+KDDAug$^\ddagger$~(10\%) & & 51.26 & 82.28 & 17.13 & 44.37 & 62.03 & 80.47 & \textbf{41.41} & 53.47 & 56.13\\
                ~~+KDDAug$^+$$^\ddagger$~(10\%) & \boldcheckmark & \textbf{52.53} & \textbf{83.16} & \textbf{17.76} & \textbf{46.01} & \textbf{62.71} & \textbf{80.76} & 41.34 & \textbf{54.63} & \textbf{57.17} \\
    			\hline
    		\end{tabular}
    	}
	\end{center}
	\caption[]{Accuracies (\%) on VQA-CP v2 and VQA v2. $^\ddagger$ denotes using $\mathcal{D}_{\text{aug}^{-}}^{\text{basic}}$. ``\textbf{Extra}" denotes using $\mathcal{D}^{\text{extra}}_{\text{aug}}$.}
	\label{tab:extend_all_qtype}
\end{table*}

\textbf{Effects of Different Teachers (Q4).} 
To show the effectiveness of multi-teacher KD strategy, we compared three teachers with different ID teacher weight $w_{id}$ and OOD teacher weight $w_{ood}$: 1) \emph{Averaged Teacher} (Simple Avg.): $w_{id} = w_{ood} = 0.5$. 2) \emph{Only ID Teacher} (ID-distill.): $w_{id} = 1, w_{ood} = 0$. 3) \emph{Only OOD Teacher} (OOD-distill.): $w_{id} = 0, w_{ood} = 1$. All results are shown in Table~\ref{tab:teacher_effectiveness}.

\textbf{\emph{Results for Q4.}} From Table~\ref{tab:teacher_effectiveness}, we have several observations: 1) Learning from all teachers can improve the performance over baseline. 2) Learning from fixed-weight teachers (\ie, single-teacher KD) can't achieve good performance on both ID and OOD settings simultaneously, \eg, ``OOD-distill." obtains OOD performance gains (+18.66\%) while suffering from a signiﬁcant drop on ID performance (-1.98\%). 3) In contrast, our dynamic multi-teacher strategy increases OOD performance by 20.50\% while ID performance drops slightly by 0.62\%, and achieves the best trade-off performance, which proves its effectiveness.

\textbf{Effects of Augmentation Diversity (Q5).}
To explore the effects of more diverse augmentation types, we compared KDDAug and KDDAug$^+$ with $D^{\text{basic}}_{\text{aug}}$ and $D^{\text{extra}}_{\text{aug}}$. For fair comparison, we didn't use initial answers and removed all paraphrasing samples from $D^{\text{basic}}_{\text{aug}}$~(denoted as $D^{\text{basic}}_{\text{aug}^-}$) since they are a subset of $\mathcal{D}_{extra}$. Meanwhile, we sampled same number of samples of the ``\texttt{Other}" cateogry samples with $D^{\text{basic}}_{\text{aug}^-}$ from $D^{\text{extra}}_{\text{aug}}$\footnote{$D^{\text{basic}}_{\text{aug}^-}$ and $D^{\text{extra}}_{\text{aug}}$ have same \Qyesno~and \Qnumber~category samples.}. We compared them on different size of samples~(different $\alpha$ values in CLIP-based filtering). Results are shown in Table~\ref{tab:extend_all_qtype}.

\textit{\textbf{Results for Q5.}} From the results, we can observe that with more diverse augmented samples, KDDAug can consistently improve both ID and OOD performance for all different $\alpha$, especially on ``\texttt{Other}" category (\eg, $> 0.78\%$ gains). In particular, even if we don't rely on any rule-based answers, KDDAug surpasses the baseline UpDn model on all categories on VQA-CP v2 when $\alpha = 50$ or 100.

\subsection{Visualization Results}
We show some augmented samples by KDDAug in Fig.~\ref{fig:visualization}. From Fig.\ref{fig:visualization}, we can observe that our KDDAug can compose potential reasonable VQ pairs and assign satisfactory pseudo labels. Take the third question ``\texttt{What garnish is on the pizza?}" as an example, there are multiple garnishes on the pizza, and KDDAug cleverly assigns multiple answers: ``\texttt{pepperoni}", ``\texttt{peppers}" and ``\texttt{cheese}", which demonstrates the superiority of the ``soft" pseudo labels generated by KDDAug.

\section{Conclusions and Future Work}

In this paper, we proposed a model-agnostic Knowledge Distillation based Data Augmentation (KDDAug) for VQA. KDDAug relaxes the requirements for pairing reasonable image-question pairs, and utilizes a multi-teacher KD to generate robust pseudo labels for augmented samples. KDDAug can consistently improve both ID and OOD performance of different VQA baselines. We validated the effectiveness of KDDAug through extensive experiments. Moving forward, we are going to 1) extend the KDDAug-like DA strategy to other visual-language tasks (\eg, captioning~\cite{chen2017sca,chen2021human,mao2022rethinking} or grounding~\cite{chen2021ref,chen2020rethinking,lu2019debug,xiao2021boundary}); 2) design some specific training objectives (\eg, contrastive loss) to further benefit from these augmented samples. 3) further improve the generalization ability of VQA models by incorporating other available large-scale datasets.

\noindent\textbf{Acknowledgement.}
This work was supported by the National Key Research \& Development Project of China (2021ZD0110700), the National Natural Science Foundation of China (U19B2043, 61976185), Zhejiang Natural Science Foundation (LR19F020002), Zhejiang Innovation Foundation(2019R52002), and the Fundamental Research Funds for the Central Universities (226-2022-00087).

\bibliographystyle{splncs04}
\bibliography{egbib}

\appendix
\title{Rethinking Data Augmentation for Robust Visual Question Answering \\
****** Supplementary Manuscript ******}

\author{Long Chen\inst{1}\thanks{Long Chen and Yuhang Zheng are co-first authors with equal contribution.} \and
Yuhang Zheng\inst{2}$^\ast$ \and
Jun Xiao\inst{2}\thanks{Corresponding author. Codes: \url{https://github.com/ItemZheng/KDDAug}.}}
\authorrunning{L. Chen and Y. Zheng et al.}
\institute{$^1$Columbia University \qquad
$^2$Zhejiang University \\
\email{zjuchenlong@gmail.com, itemzhang@zju.edu.cn, junx@cs.zju.edu.cn}
}

\maketitle

This supplementary manuscript is organized as follows:

\begin{enumerate}

    \item In Section~\ref{sec:a}, we introduce more details about experiments, including experimental settings, training process, KDDAug settings, and paraphrasing steps mentioned in Section~\ref{sec:4.1} (\cf, Experimental Settings and Implementation Details).
    
    \item In Section~\ref{sec:b}, we describe more details about the CLIP$_{\text{rank}}$ used in Section~\ref{sec:4.4} (\cf, ablation studies Q1 \& Q2).
    
    \item In Section~\ref{sec:c}, we add additional experimental results to demonstrate the effects of augmentation diversity.
    
    \item In Section~\ref{sec:d}, we demonstrate more visualization results, including comparisons of the generated pseudo ground-truth answers between KDDAug and SimpleAug~\cite{kil2021discovering}, and more diversity augmented samples in $D^{\text{extra}}_{\text{aug}}$.

\end{enumerate}

\section{More Details about Experiments}\label{sec:a}

\subsection{Details about Experimental Settings}
\noindent\textbf{Datasets.} We evaluated the proposed KDDAug on two datasets: the ID benchmark \textbf{VQA v2}~\cite{goyal2017making} and OOD benchmark \textbf{VQA-CP v2}~\cite{agrawal2018don}. VQA v2 is a ``balanced" VQA dataset, where each question has complementary images with opposite answers. Although VQA v2 has reduced language biases to some extent, the statistical biases from questions still can be leveraged~\cite{agrawal2018don}. To disentangle the biases and clearly monitor the progress of VQA, VQA-CP re-organizes VQA v2, and deliberately keeps different QA distributions in the training and test sets.

\noindent\textbf{Evaluation Metrics.} For model accuracies, we followed standard VQA evaluation metric~\cite{antol2015vqa}, and reported accuracy on three different categories separately: Yes/No (\texttt{Y/N}), number counting (\texttt{Num}), and other (\texttt{Other}) categories. For the ID evaluation, we reported the results on the VQA v2 val set. For the OOD evaluation, we reported the results on the VQA-CP v2 test set. Meanwhile, we followed~\cite{niu2021counterfactual} and used Harmonic Mean~(\textbf{HM}) of the accuracies on both two datasets (VQA v2 val \& VQA-CP test) to evaluate the trade-off between ID and OOD evaluations.

\subsection{Details about Training Process}
To effectively train VQA models with both original and new augmented samples, we first pre-trained VQA models with only original samples following their respective settings. Then, we fine-tuned pre-trained VQA models\footnote{We only fine-tune the basic VQA backbone (UpDn) in the fine-tuning stage, \ie, for ensemble-based models, we removed the auxiliary question-only branches. \label{footnote: question-only}} with augmented samples for 5 epochs. The batch size was set to 512. We used the Adamax~\cite{diederik2015adam} as the optimizer and the random seed was set to 0.

\subsection{Details about KDDAug Settings}
For the object detector, we used the Faster R-CNN~\cite{ren2015faster} pre-trained on VG~\cite{krishna2017visual} to detect 36 objects and attributes (\eg, color) for every image. To keep highly-confident predictions, we set the score thresholds for object and attribute to 0.8 and 0.4. \emph{It is worth noting that we only used detection results from the Faster R-CNN without relying on other extra human annotations.}

\subsection{Details about Paraphrasing} 

In this section, we introduce more details about the paraphrasing in Section~\ref{sec:4.1}. Paraphrasing is a supplementary data augmentation trick proposed by SimpleAug~\cite{kil2021discovering} which composes new VQ pairs by searching similar questions. Specifically, for each original sample~($I_i, Q_i, a_i$), if question $Q_j$ is similar to $Q_i$, they construct a new augmented training sample~($I_i, Q_j, a_i$). By ``similar", we mean that the cosine similarity between question BERT embeddings\footnote{The BERT is pre-trained on BookCorpus~\cite{zhu2015aligning} and English Wikipedia. We get the pre-trained BERT model from \url{https://github.com/google-research/bert}.}~\cite{devlin2019bert} is large than $0.95$. For each original sample, we choose all top-3 similar questions for composing new samples according to cosine similarity scores. 

\section{Details about CLIP$_{\text{rank}}$} \label{sec:b}

CLIP$_{\text{rank}}$ aims to rank the quality of all SimpleAug assigned answers, \ie, we used it to rank the similarity between each image and the augmented question-answer (QA) pair. We firstly generated a prompt for each augmented QA pair, and utilized a pretrained CLIP~\cite{radford2021learning} to calculate the similarity score between the prompt and the image. Specifically, we designed different strategies to generate prompts for different question types. For \Qnumber~and \Qcolor~questions, we generated prompts by removing the question type category prefix and inserting the answer in front of the noun\footnote{For \Qnumber~and \Qcolor~questions, there is an only single noun in the questions.}. For example, if the question is ``\texttt{how many umbrellas are there}", and its pseudo answer is ``\texttt{2}", then the prompt is ``\texttt{2 umbrellas are there}". For the other questions, we generated prompts by simply replacing question type category prefix with the answer. For example, if the question is ``\texttt{what food is that}", and its pseudo answer is ``\texttt{donut}", then its prompt is ``\texttt{donut is that}". Based on their respective similarity scores, we ranked all the augmented samples.

\setcounter{figure}{5}
\section{Additional experimental results} \label{sec:c}
\setcounter{table}{7}
\begin{table*}[t]
	\begin{center}
		\begin{tabular}{|l | c | c | c |c|}
			\hline
			\textbf{Models} & \textbf{Extra} & VQA-CP v2 & VQA v2 & \textbf{HM} \\
		    \hline
		    KDDAug$^\ddagger$ & & 53.03 & 61.59 & 59.99 \\
		    ~~+Initial Answers & & 59.02 & 61.07 & 60.03\\
            \hline
            KDDAug$^{+\ddagger}$ & \boldcheckmark & 53.76 & 62.58 & 57.84\\
            ~~+Initial Answers& \boldcheckmark& \textbf{60.03} & \textbf{62.24} & \textbf{61.12} \\
			\hline
		\end{tabular}
	\end{center}
	\caption[]{Accuracies (\%) on VQA-CP v2 and VQA v2.$^\ddagger$ denotes without paraphrasing samples. \textbf{``Extra"} denotes using $D^{extra}_{aug}$.}
	\label{tab:extend_all_qtype_2}
\end{table*}

In this section, we add additional experimental results to demonstrate the effects of augmentation diversity. As shown in Table~\ref{tab:extend_all_qtype_2}, diversity indeed helps model performance (both w/ and w/o initial answers.). However, when using the extra paraphrasing~\cite{kil2021discovering}, the improvement gains brought by diversity in the smaller size KDDAug$^{+\ddagger}$ is overwhelmed. Moreover, we only use KDDAug for SOTA comparison rather than KDDAug$^+$ for two reasons: 1) The sample size of whole $D_{\text{aug}}^{\text{extra}}$ is enormous. Thus, it is infeasible to directly train models with whole $D_{\text{aug}}^{\text{extra}}$ (KDDAug$^+$). 2) For efficiency and fair comparison with prior work SimpleAug~\cite{kil2021discovering}, we controlled the number of samples to be the same as SimpleAug (\ie, KDDAug$^{+\ddagger}$ in Table~\ref{tab:extend_all_qtype} and \ref{tab:extend_all_qtype_2}).

\section{More Visualization Results} \label{sec:d}

\begin{figure}[t]
	\centering
	\includegraphics[width=\linewidth]{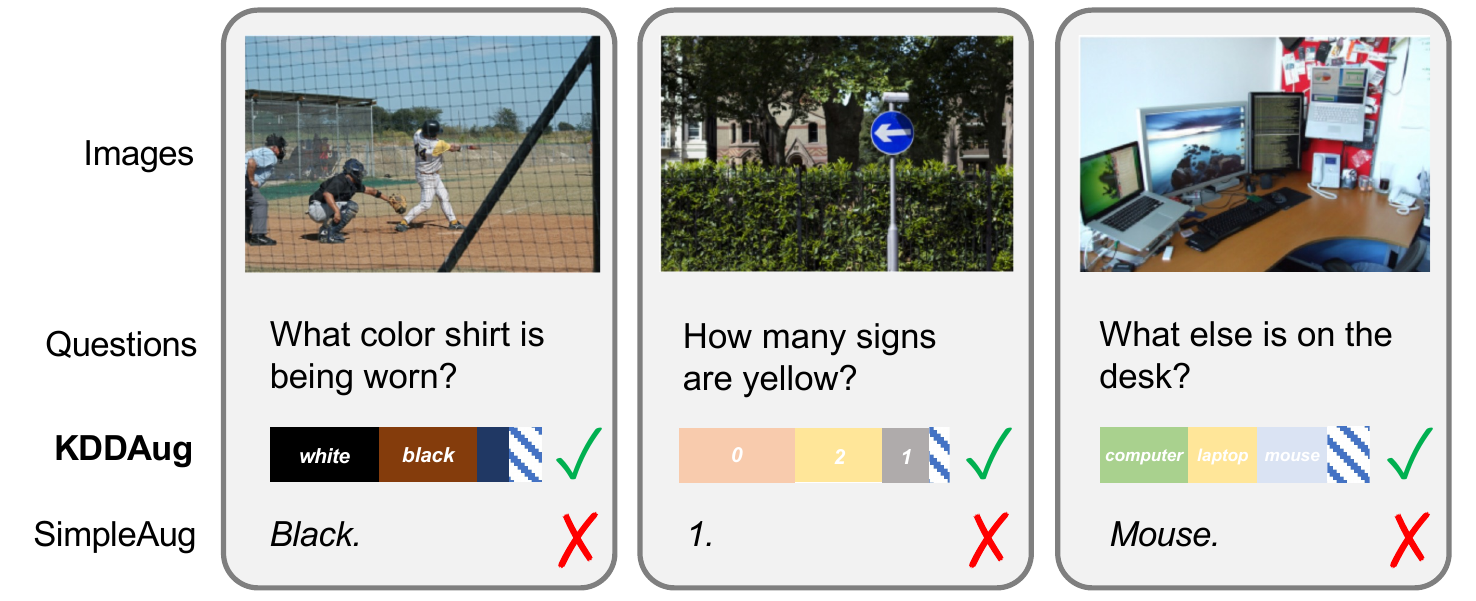}
	\caption{Visualization results of some augmented samples and their pseudo ground-truth answers assigned by our KDDAug and SimpleAug.}
	\label{fig:visualization_1}
\end{figure}

\begin{figure}[t]
	\centering
	\includegraphics[width=\linewidth]{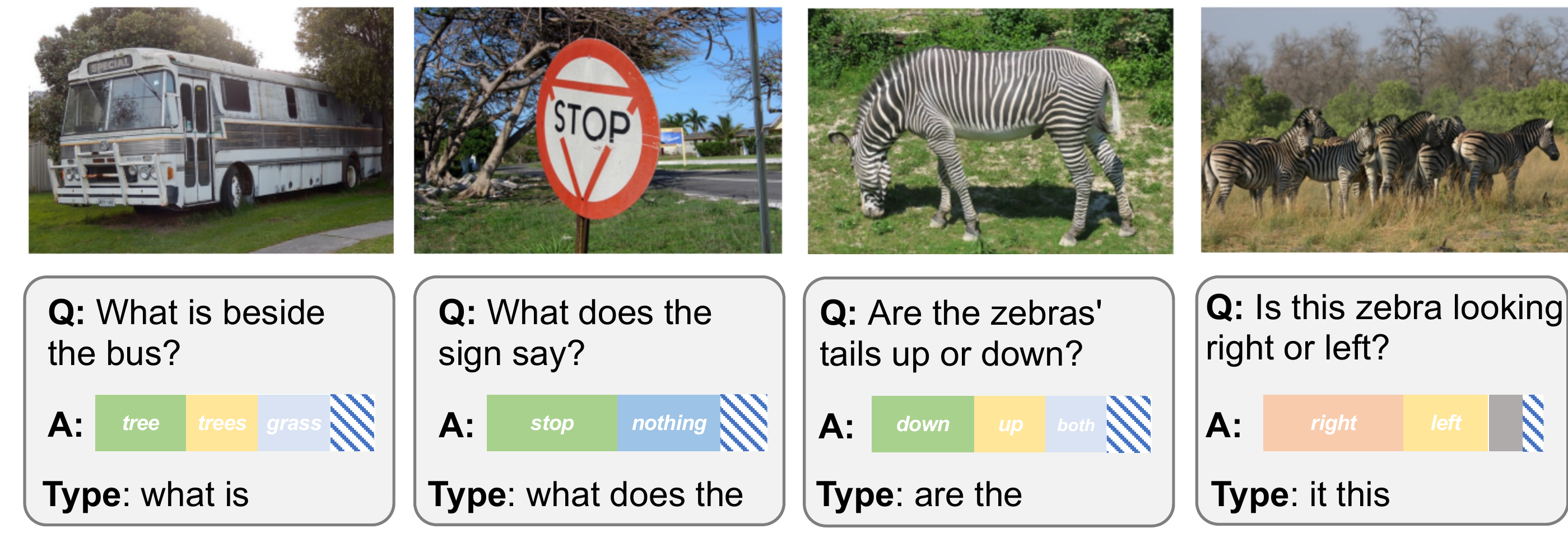}
	\caption{Visualization results of some augmented samples in $D^{extra}_{aug}$. ``Type" denotes the question type, which are all excluded in original SimpleAug.}
	\label{fig:visualization_2}
\end{figure}

\subsection{KDDAug vs. SimpleAug}
To further compare KDDAug and SimpleAug, we show some augmented samples and their answers assigned by KDDAug and SimpleAug in Fig.~\ref{fig:visualization_1}. Take the second question ``\texttt{How many signs are yellow?}" as an example, SimpleAug directly uses the count of signs appearing in the image as the answer, \eg, ``\texttt{1}". In contrast, our KDDAug takes ``\texttt{0}" as the answer, which demonstrates the robustness of KDDAug assigned answers. Meanwhile, for some questions with multiple possible answers (\eg, the question ``\texttt{what else is on the desk}" for the third sample), our ``soft" version ground-truth answer is inherently more accurate and better for VQA model training.

\subsection{Augmented Samples in $D^{\text{extra}}_{\text{aug}}$}
As shown in Fig.~\ref{fig:visualization_2}, we show some augmented samples in $D^{\text{extra}}_{\text{aug}}$. All these samples can't be generated by SimpleAug due to the limitations of its image-question pair composition strategy. Take the third question ``\texttt{Are the zebras' tails up or down?}" as an example, SimpleAug can't generate it since it doesn't belong to \Qcolor, \Qwhat, \Qnumber~or \Qyesno~questions. In contrast, our KDDAug can generate this augmented sample and assign a reasonable answer ``\texttt{down}" for it, which demonstrates the generalization of KDDAug.

\end{document}